\documentclass[a4paper,twoside]{article}

\usepackage{epsfig}
\usepackage{subfigure}
\usepackage{calc}
\usepackage{amssymb}
\usepackage{amstext}
\usepackage{amsmath}
\usepackage{amsthm}
\usepackage{multicol}
\usepackage{pslatex}
\usepackage{apalike}
\usepackage{algorithm}
\usepackage{algpseudocode}
\usepackage{bm}
\usepackage{epstopdf}
\usepackage{siunitx}
\usepackage{balance}
\usepackage{subfigure} 
\usepackage{SCITEPRESS}     

\newcommand{\tabref}[1]{Table~\ref{#1}}
\newcommand{\figref}[1]{Figure~\ref{#1}}
\renewcommand{\eqref}[1]{Equation~(\ref{#1})}
\renewcommand{\algref}[1]{Algorithm~(\ref{#1})}

\begin{document}

\title{Cataglyphis ant navigation strategies solve the global localization problem in robots with binary sensors}

\author{\authorname{Nils Rottmann\sup{1}, Ralf Bruder\sup{1}, Achim Schweikard\sup{1} and Elmar Rueckert\sup{1}}
\affiliation{\sup{1}Institute for Robotics and Cognitive Systems, University of Luebeck, Ratzeburger Allee 160, 23562 Luebeck, Germany}
\email{\{rottmann,bruder,schweikard,rueckert\}@rob.uni-luebeck.de}
} 


\abstract{Low cost robots, such as vacuum cleaners or lawn
mowers, employ simplistic and often random navigation policies.
Although a large number of sophisticated localization and planning
approaches exist, they require additional sensors like LIDAR
sensors, cameras or time of flight sensors. In this work, we propose a global localization method biologically inspired by simple insects, such as the ant Cataglyphis that is able to return from distant locations to its nest in the desert without any or with limited perceptual cues. Like in Cataglyphis, the underlying idea of our localization approach is to first compute a pose estimate from pro-prioceptual sensors only, using land navigation, and thereafter refine the estimate through a systematic search in a particle filter that integrates the rare visual feedback.\\
In simulation experiments in multiple environments, we demonstrated that this bioinspired principle can be used to compute accurate pose estimates from binary visual cues only. Such intelligent localization strategies can improve the performance of any robot with limited sensing capabilities such as household robots or toys.}
\onecolumn \maketitle \normalsize \vfill

\section{\uppercase{Introduction}}
\label{sec:introduction}

\noindent Precise navigation and localization abilities are crucial for animal and robots. These elementary skills are far developed in complex animals like mammals \cite{o1978hippocampus}, \cite{redish1999beyond}. Experimental studies have shown that transient neural firing patterns in place cells in the hippocampus can predict the animals location or even complete future routes \cite{pfeiffer2013hippocampal}, \cite{foster2006reverse}, \cite{johnson2007neural}, \cite{carr2011hippocampal}. These neural findings have inspired many computational models based on attractor networks \cite{hopfield1982neural}, \cite{samsonovich1997path}, \cite{mcnaughton2006path}, \cite{erdem2012goal} and have been successfully applied to humanoid robot motion planning tasks \cite{Rueckert2016a}, \cite{Tanneberg2019}. \\

\begin{figure}[t]
\subfigure[''Cataglyphis nodus'' by \cite{Christof1Bobzin}.]{\includegraphics[width=0.21\textwidth]{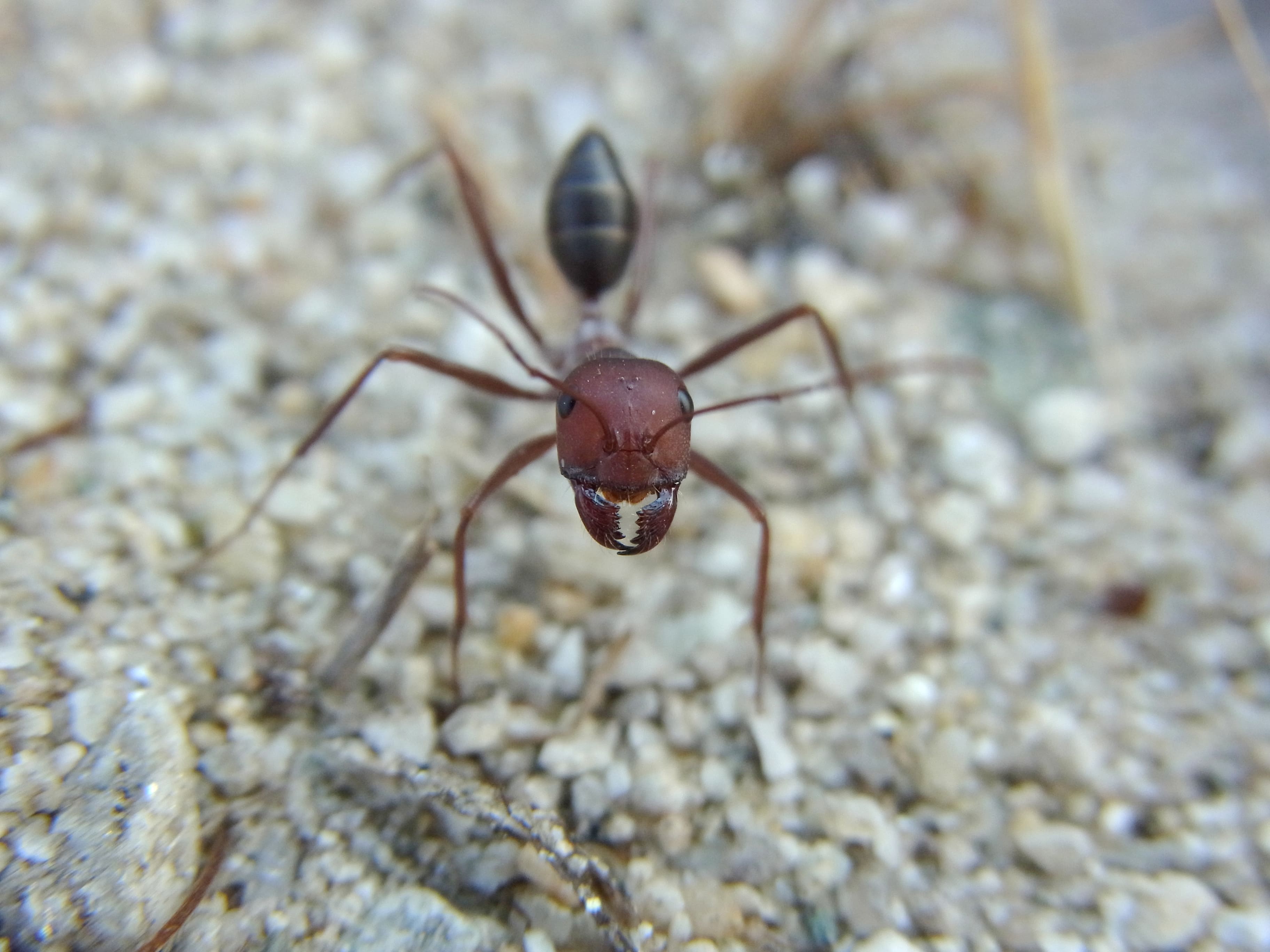}}
\hfill
\subfigure[Create 2]{\includegraphics[width=0.237\textwidth]{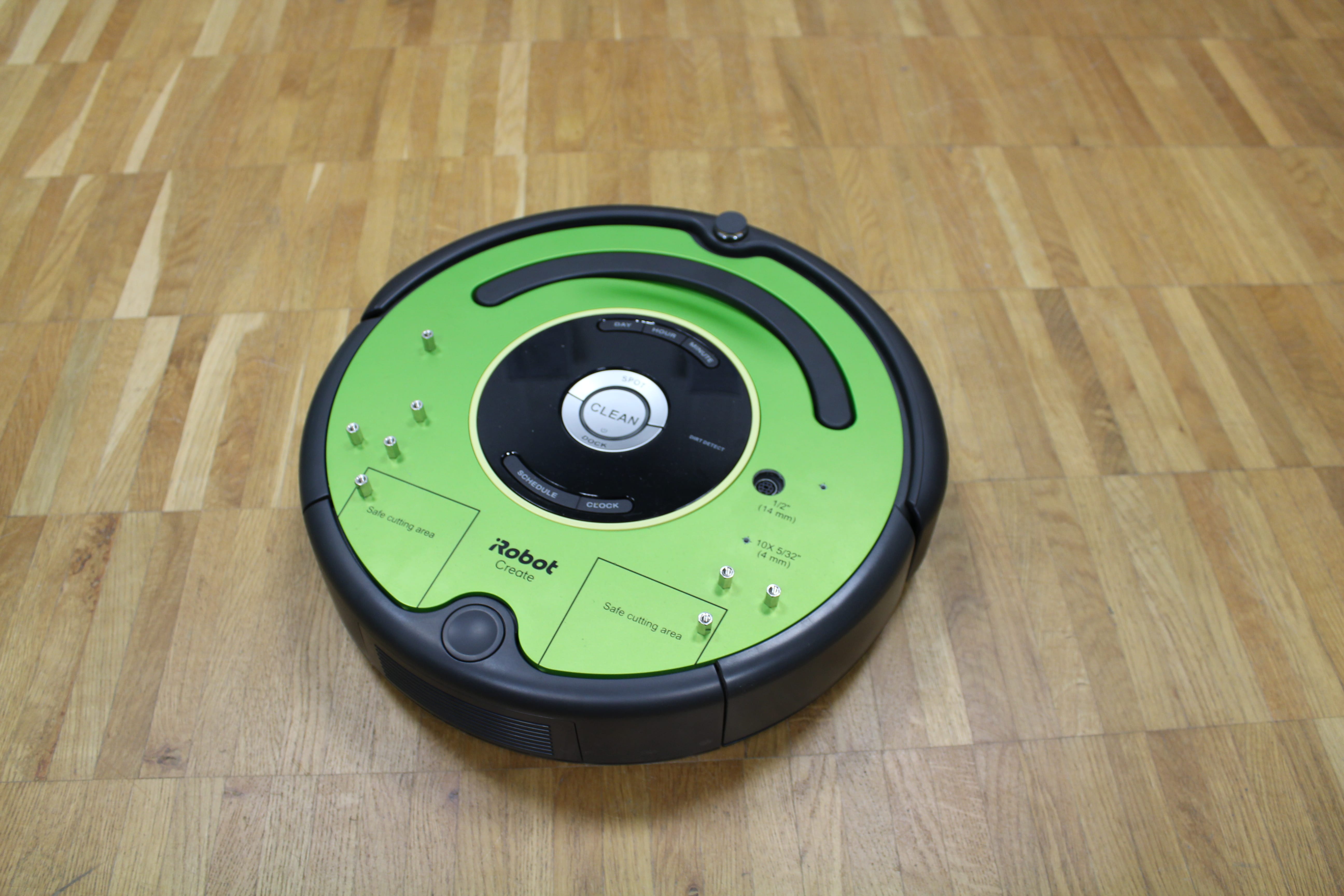}}
\caption{The Cataglyphis ant and the robot Create 2 from the company iRobot.}
\label{fig:AntAndRobot}
\end{figure}

\noindent In this work we took inspiration from simpler insects which are able to precisely navigate by using sparse perceptual and pro-prioceptual sensory information. For example, the desert ant Cataglyphis (\figref{fig:AntAndRobot}a) employs basic path integration, visual piloting and systematic search strategies to navigate back to its nest from distant locations several hundred meters away \cite{wehner1987spatial}. While computational models of such navigation skills lead to better understanding of the neural implementation in the insect \cite{lambrinos2000mobile}, they also have a strong impact on a large number of practical robotic applications. Especially, non-industrial autonomous systems like household robots (\figref{fig:AntAndRobot}b) or robotic toys require precise navigation features utilizing low-cost sensory hardware \cite{lee2012balancing}, \cite{miller1995design}, \cite{nebot1999initial}. \\

\noindent Despite this obvious need for precise localization in low cost systems, most related work in mobile robot navigation can be categorized into two \textit{extreme} cases. Either computational and monetary expensive sensors like cameras or laser range finders are installed \cite{su2017global}, \cite{feng2017global}, \cite{ito2014w} or simplistic navigation strategies such as a random walks are used like in the current purchasable household robots \cite{tribelhorn2007evaluating}. When using limited sensing, only few navigation strategies have been proposed. O'Kane et al. investigated how complex a sensor system of a robot really has to be in order to localize itself. Therefore, they used a minimalist approach with contact sensors, a compass, angular and linear odometers in three different sensor configurations \cite{o2007localization}. Erickson et al. addressed the localization problem for a blind robot with only a clock and a contact sensor. They used probabilistic techniques, where they discretized the boundary of the environment into small cells. A probability $P_{k,i}$ of the robot being in the cell $i$ at the time step $k$ is allocated to each one and updated in every iteration. For active localization, they proposed an entropy based approach in order to determine uncertainty-reducing motions \cite{erickson2008probabilistic}. Stavrou and Panayiotou on the other hand proposed a localization method based on Monte Carlo Localization \cite{dellaert1999monte} using only a single short-range sensor in a fixed position \cite{stavrou2012localization}. The open challenge however that remains is how to efficiently localize a robot equipped only with a single binary sensor and odometer. Such tasks arise, for example, especially with autonomous lawn mowing robots. They usually use a wire signal to detect whether they are on the area assigned to them or not. There are also sensors that detect the moisture on the surface and can thus detect grass \cite{bernini2009lawn}. All these sensors usually return a binary signal indicating whether the sensor is in the field or outside. The aforementioned approaches can either not be applied to such localization tasks because they require low-range sensors \cite{stavrou2012localization} or they are trying to solve the problem with even less sensor information \cite{erickson2008probabilistic}. A direct comparison between these approaches and our method is not possible since different sensors types are used. However, to be able to compare the estimation accuracy of our methods with the aforementioned algorithms, we created our test environments similar to the test envrionment used in \cite{stavrou2012localization}.\\

\noindent We propose an efficient and simple method for global localization based only on odometry data and binary measurements which can be used in real time on a low-cost robot with only limited computational resources. This work demonstrates for the first time how basic navigation principles of the Cataglyphis ant can be implemented in an autonomous outdoor robot. Our approach has the potential to replace the currently installed naive random walk behavior in most low cost robots to an intelligent localization and planning strategy inspired by insects. The methods we used are discussed in Section II. We first present a wall following scheme which is required to steer the robot along the boundary line of the map based on the binary measurement data. Inspired by path integration we determine dominant points (DPs) to represent the path estimated by the odometry. We then generate a first estimate of the pose of the robot using land navigation. Therefore, we compare the shape of the estimated path with the given boundary line. The generated pose estimate allows to systematically search for the true pose in the given area using a particle filter. This leads to a more accurate localization of the robot. Here a particle filter is required since other localization techniques, such as Kalman Filter, are not able to handle binary measurement data. In Section III, we evaluate our approach in two simulated environments and show that it leads to accurate pose estimates of the robot. We conclude in Section IV.

\section{\uppercase{Methods}}

\begin{figure}[btp]
\centering
\includegraphics[width=0.43\textwidth]{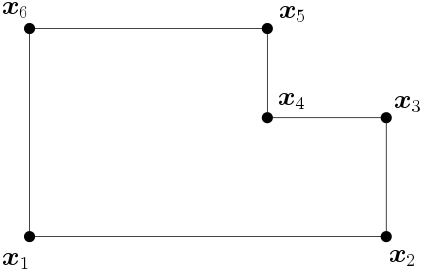}
\caption{Boundary Map: An example map of an environment which defines the boundary by connecting vertices ${X = [\boldsymbol{x}_1, \boldsymbol{x}_2, ..., \boldsymbol{x}_6]}$ to a polygon shape. }
\label{fig:PolygonMap}
\end{figure}

\begin{figure}[b]
\centering
\includegraphics[width=0.43\textwidth]{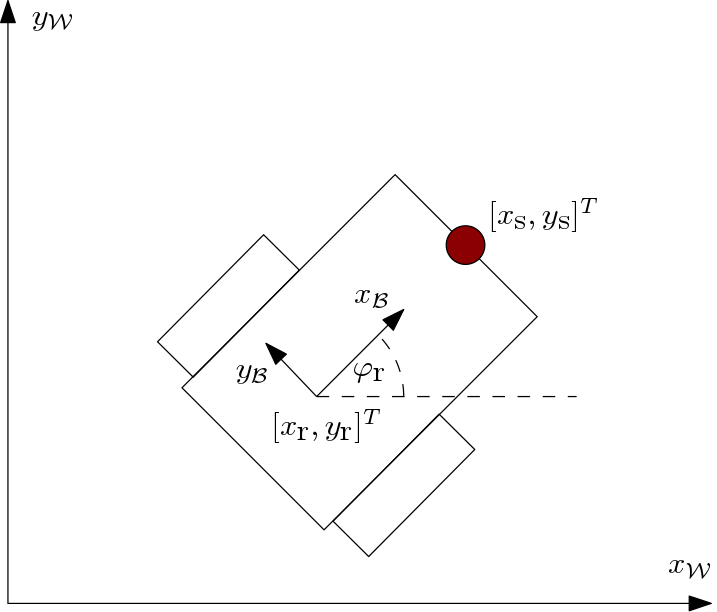}
\caption{A differential drive robot within the ${2 \text{-dimensional}}$ world frame given by $\mathcal{W}$. The robots main frame $\mathcal{B}$ is determined by the frame for the odometry. The robots position is given by $x_{\text{r}}$, $y_{\text{r}}$ and the sensor position by $x_{\text{s}}$, $y_{\text{s}}$.}
\label{fig:DiffDrive}
\end{figure}

\noindent For our proposed localization method, we assume that a map of the environment is given as a boundary map $\mathcal{M}$ defined as a polygon with vertices ${X = [\boldsymbol{x}_1, \boldsymbol{x}_2, ..., \boldsymbol{x}_n]}$ with $\boldsymbol{x}_i \in \mathbb{R}^2, \, \forall i \in [1,n]$. In \figref{fig:PolygonMap} such a boundary map is shown. As sensor signals we only receive the binary information $s \in \{0,1\}$, where $0$ means that the sensor is outside of the map and $1$ means the sensor is within the boundary defined by the polygon. We are using a simple differential drive robot controlled by the desired linear and angular velocities $v$ and $\omega$. Furthermore, odometry information is assumed to be given. We assume the transform between the sensor position and the odometry frame, which is also the main frame of the robot and has the position $[x_{\text{r}},y_{\text{r}}]^{\top}$ in world coordinates, is known as
\begin{equation}
\label{eq:SensorPosition}
\begin{bmatrix}
x_{\text{s}} \\
y_{\text{s}}
\end{bmatrix} =
\begin{bmatrix}
x_{\text{r}} \\
y_{\text{r}}
\end{bmatrix} + 
\boldsymbol{R}(\varphi_{\text{r}})
\begin{bmatrix}
\Delta x \\
\Delta y
\end{bmatrix} .
\end{equation}
Here, $\boldsymbol{R}(\varphi_{\text{r}})$ is the two dimensional rotation matrix with the orientation angle $\varphi_{\text{r}}$ of the robot and $\Delta x$, $\Delta y$ the distances between odometry frame and sensor position in robot coordinates. In \figref{fig:DiffDrive} the setup for the differential drive robot is depicted. For our proposed localization method, it is required that there is a lever arm between the robots frame and the sensor, such that if only the orientation of the robot changes the position of the sensor changes too. 

\subsection{Wall Follower}

\begin{figure}[btp]
\centering
\includegraphics[width=0.43\textwidth]{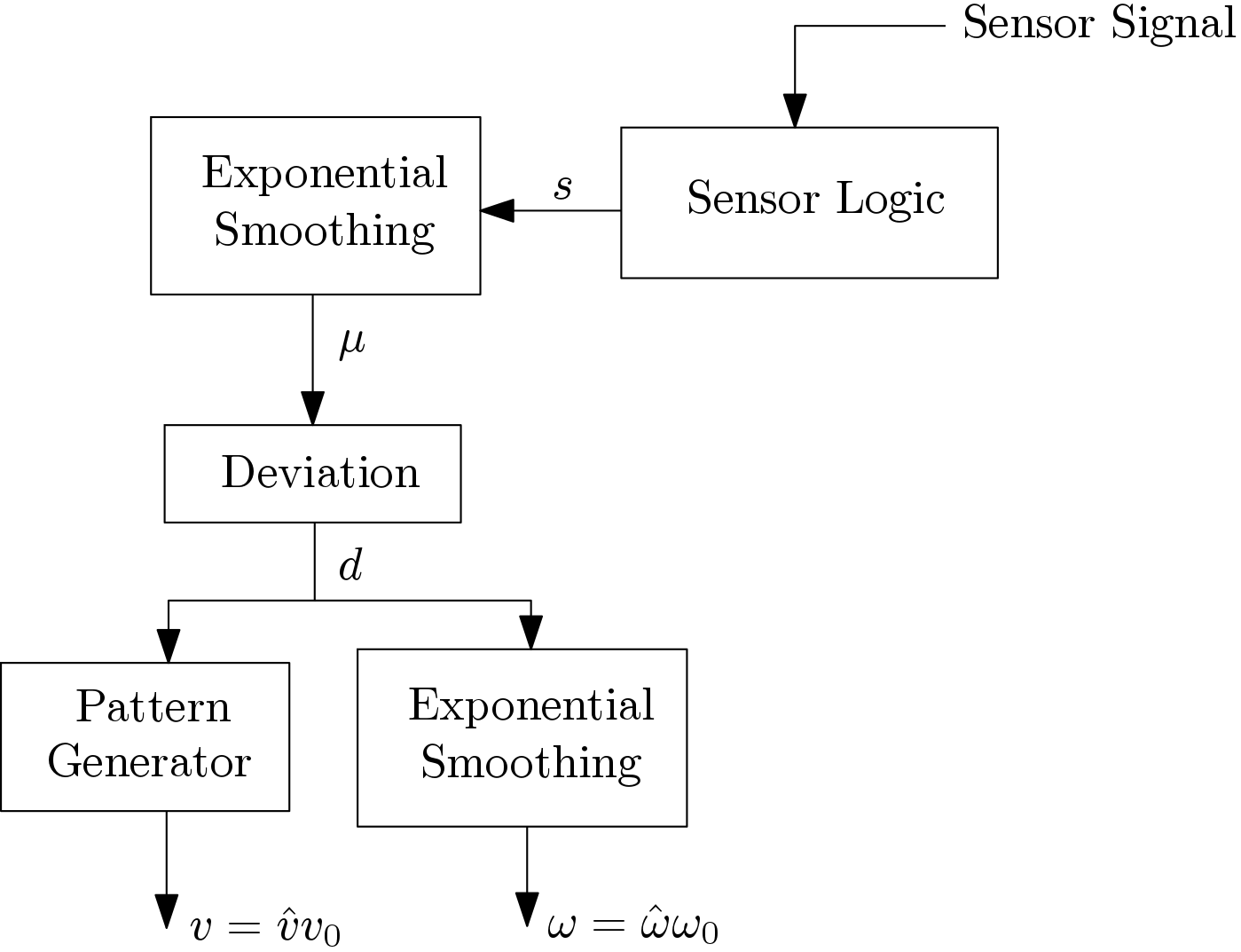}
\caption{Wall Follower: Control method blocks representing the wall following algorithm.}
\label{fig:WallFollower}
\end{figure}

Since only at the boundary we are able to generate sensor data useful for localization, a wall follower is required. The proposed control algorithm allows to navigate the robot along the boundary line even if the binary measurements are corrupted by noise. Thereby, to increase the scanning area, the robot moves in wiggly lines along the boundary such that the mean of the sensor detection is ${\mu_{\text{d}} = 0.5}$. For the update step of the sensor mean we use exponential smoothing
\begin{equation}
\label{eq:amu}
\mu_{\text{d}} \leftarrow a_{\mu} \mu_{\text{d}} + (1 - a_{\mu}) s ,
\end{equation}
starting with an initial mean of $\mu_{\text{d}} = 1.0$ (assumption: start within the field). Here $a_{\mu}$ defines the update rate of the mean $\mu_{\text{d}}$. We then calculate the difference between the desired mean and the actual mean
\begin{equation}
\label{eq:difference}
d = 2 \left(\frac{1}{2} - \mu_{\text{d}}\right) .
\end{equation}
The difference is scaled onto the range $[-1,1]$. We now use relative linear and angular velocities $\hat{v}$ and $\hat{\omega}$ for the control algorithm. They have to be multiplied with the maximum allowed velocities for the given robotic system $v_0$ and $\omega_0$. Using the difference $d$ from \eqref{eq:difference}, we update the relative forward velocity for the differential drive robot $\hat{v}$ using again exponential smoothing
\begin{equation}
\label{eq:av}
\hat{v} \leftarrow a_v \hat{v} + (1 - a_v) (1 - |d|) . 
\end{equation}
Here, $a_v$ defines an update rate and $|d|$ the absolute value of $d$. The relative velocity increases the longer and better the desired mean of ${\mu_{\text{d}} = 0.5}$ is reached. Thereby, the relative velocity is always in the range of ${\hat{v} \in [0,1]}$. The relative angular velocity of the robot is determined using the deviation $d$ to the desired mean and a stabilizing term ${\cos\left(2 \pi \frac{c}{M}\right)}$ to ensure robustness against noise corrupted measurements
\begin{equation}
\hat{\omega} = \frac{1}{2} \left(d + \cos\left(2 \pi \frac{k}{K}\right) \right) .
\end{equation}
Here $k$ is a counter variable and $K$ the counter divider. Thus, $K$ should be chosen accordingly to the frequency of the operating system. A diagram of the controller as well as pseudo code can be found in \figref{fig:WallFollower} and \algref{alg:WallFollower} respectively.

\begin{algorithm}[tb]
\caption{Wall Follower}
\label{alg:WallFollower}
\begin{itemize}
\item \textbf{Parameters} \\
		$K$, $a_{\mu}$, $a_v$, $v_0$, $\omega_0$
\item \textbf{Inputs} \\
		$s$
\item \textbf{Outputs} \\
		$v$, $\omega$
\end{itemize}
\begin{algorithmic}[1]
\State $k = 0$
\State $\mu_d = 1.0$ \Comment{start within the field}
\While {true}
\If {Mode == 0} \Comment{search for boundary}
	\State $\mu_d \leftarrow a_{\mu} \mu_d + (1-a_{\mu}) s$
	\If {$\mu_d > 0.5$}	
		\State $v = v_0$
		\State $\omega = 0$
	\Else
	    \State Mode $= 1$
	\EndIf
\ElsIf {Mode == 1} \Comment{wall following}
	\State $\mu_d \leftarrow a_{\mu} \mu_d + (1-a_{\mu}) s$
	\State $d = 2 \cdot (0.5 - \mu_d)$
	\State $\hat{v} \leftarrow a_v \hat{v} + (1-a_v) (1 - ||d||)$
	\State $v = \hat{v} v_0$
	\State $\omega = \left(d + \cos\left(2 \pi \frac{\text{k}}{K}\right)\right) \cdot 0.5 \cdot \omega_0$	
\EndIf
\State $k \leftarrow k + 1$
\State \Return{$v$, $\omega$}
\EndWhile
\end{algorithmic}
\end{algorithm}

\subsection{Path Integration / Land Navigation}

In order to get a first pose estimate for the robot we use land navigation between the given map $\mathcal{M}$ and the driven path $\mathcal{P}$ along the boundary line. \\

\noindent First, we start by generating a piecewise linear function of the orientation of the boundary line in regard to the length ${\theta_{\text{b}} = \theta_{\text{b}}(x)}$. This function represents the shape of the boundary and is defined as
\begin{equation}
\theta_{\text{b}}(x) = \phi_i \quad \text{for} \quad l_i \leq x < l_{i+1}, i=1,2,\dots,2n .
\end{equation}
with
\begin{equation}
\begin{split}
\phi_{i+1} &= \phi_i + \Delta \theta_i \\
l_{i+1} &= l_i + ||\boldsymbol{v}_i|| .
\end{split}
\end{equation}
Here, $||\boldsymbol{v}_i||$ defines the Euclidean norm for the vector $\boldsymbol{v}_i$. We start with ${\phi_1 = 0}$ and ${l_1 = 0}$. Thereby, ${\boldsymbol{v}_i = \hat{X}_{i+1} - \hat{X}_i}$ with ${\hat{X} = [\boldsymbol{x}_1, \boldsymbol{x}_2, ..., \boldsymbol{x}_n, \boldsymbol{x}_1, \boldsymbol{x}_2, ..., \boldsymbol{x}_n, \boldsymbol{x}_1]}$ consisting of the vertices $X$ of the map and ${\Delta \theta_i}$ as the difference in orientation between $\boldsymbol{v}_i$ and $\boldsymbol{v}_{i-1}$.\\

\noindent Second, in order to represent the path driven by the robot as vertices connected by straight lines (similar to the map representation), we generate dominant points (DPs). Therefore, we are simply using the estimated position from the odometry of our robot starting with $DP = [\boldsymbol{x}_0]$ at the first contact with the boundary line. We also initialize a temporary set $S = [\boldsymbol{x}_0]$ which holds all the actual points which can be combined to a straight line. We start by adding in every iteration step the new position estimate from the odometry data $\boldsymbol{x}$ to our temporary set $S \leftarrow [S, \boldsymbol{x}]$ and then check if 
\begin{equation}
L_{\text{min}} > ||DP_{\text{end}} - S_{\text{end}}||
\end{equation}
and
\begin{equation}
e_{\text{max}} >  \frac{1}{N-2} \sum_{i=2}^{N-1} e_i
\end{equation}
are both true. If this is the case we add the point $S_{\text{end}-1}$ to the set of DPs and set the temporary set back to $S = [S_{\text{end}}, \boldsymbol{x}]$. Here, $e_i$ defines the shortest distance between the point $S_i$ and the vector given by $S_{\text{end}} - S_1$. Pseudo code for the algorithm can be found in \algref{alg:DPGeneration}. The parameters $L_{\text{min}}$ and $e_{\text{max}}$ are problem specific and have to be tuned accordingly.\\

\begin{algorithm}[tb]
\caption{DP Generation}
\label{alg:DPGeneration}
\begin{itemize}
\item \textbf{Parameters} \\
		$L_{\text{min}}$, $e_{\text{max}}$
\item \textbf{Inputs} \\
		$\boldsymbol{x}$
\item \textbf{Outputs} \\
		$DP$
\end{itemize}
\begin{algorithmic}[1]
\State $DP = [\boldsymbol{x}_0]$
\State $S = [\boldsymbol{x}_0]$
\If {new $\boldsymbol{x}$ available}
	\State $d = ||DP_{\text{end}} - \boldsymbol{x}||$
	\If {$d < L_{\text{min}}$}
		\State $S \leftarrow [S, \boldsymbol{x}]$
	\Else
		\State $S_{\text{tmp}} = [S, \boldsymbol{x}]$
		\State $e = \text{errorLineFit}(S_{\text{tmp}})$
		\If {$e < e_{\text{max}}$}
			\State $S \leftarrow S_{\text{tmp}}$
		\Else
			\State $DP \leftarrow [DP, S_{\text{end}}]$ 
			\State $S \leftarrow [S_{\text{end}}, \boldsymbol{x}]$
		\EndIf
	\EndIf
\EndIf
\end{algorithmic}
\end{algorithm}

\begin{figure}[btp]
\centering
\includegraphics[width=0.43\textwidth]{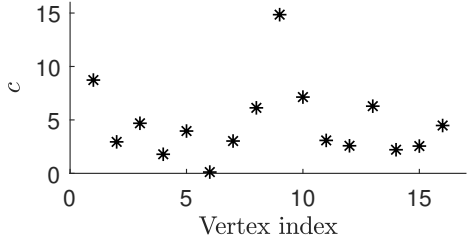}
\caption{Correlation errors between the actual DP and the vertices of the map are shown. Here the vertex with the index $6$ fulfills the required conditions and is chosen for the first position estimate.}
\label{fig:CorrelationErros}
\end{figure}

\noindent Third, every time a new DP is accumulated we generate a piecewise orientation function $\theta_{\text{r}}$ as presented above using the actual set of DPs, as long as the accumulated length between the DPs is larger then ${U_{\text{min}} \times \text{Map Circumference}}$. This ensures that the algorithm does not start too early with the comparison. The parameter $U_{\text{min}} \in [0,1]$ represents the minimal required portion of the boundary line to identify unique positions and has to be chosen or trained according to the given map. We then compare $\theta_{\text{r}}$ with $\theta_{\text{b}}$ for all vertices $\boldsymbol{x}_i, \, i=1,2,\dots,n$ of the boundary line in order to find a suitable vertex at which the robot could be. Therefore, we adjust $\theta_{\text{b}}$ such that $\theta_{\text{b},i}(x) = \theta_{\text{b}}(x + l_i) - \phi_i,\, i=2n,2n-1,\dots,n+1$. We now evaluate these functions at $N$ linearly distributed points from $-L$ to $0$ which results into vectors $\boldsymbol{\theta}_{\text{b},i}$. Here $L$ is the path length driven by the robot along the boundary line. We then calculate the correlation error
\begin{equation}
\boldsymbol{c}_i = \frac{1}{N} \sum_{k=1}^N |\boldsymbol{\theta}_{\text{b},i,k} - \boldsymbol{\theta}_{\text{r}}|,
\end{equation}
where ${\boldsymbol{\theta}_{\text{r}}}$ is the evaluation of the function ${\hat{\theta}_{\text{r}}(x) =  \theta_{\text{r}}(x + L) - \theta_{\text{r}}(L)}$ in the range from $-L$ to $0$ at $N$ linearly distributed points. We now have correlation errors between the actual driven path by the robot along the boundary line and every vertex of the polygon map. In Figure \ref{fig:CorrelationErros} correlation errors are graphically displayed. \\

\noindent Fourth, we search for the vertex with the minimal correlation error and check if the error is below a certain treshold $c_{\text{min}}$. The parameter $c_{\text{min}}$ has to be trained accordingly to the given map. If we do not find a convenient vertex, we simply drive further along the wall until finding a vertex which follows the condition above. Thereby we limit the stored path from the robot to the length of the circumference of the map. If a convenient vertex is found, we call this vertex from now on matching vertex, the first guess of the robots position can be determined as $\boldsymbol{x}_{\text{est}}$, where $\boldsymbol{x}_{\text{est}}$ is the matching vertex. Moreover, we can also define a orientation estimate as
\begin{equation}
\varphi_{\text{est}} = \text{atan2}(\boldsymbol{v}_{\text{est},y},\boldsymbol{v}_{\text{est},x}) ,
\end{equation}
where $\boldsymbol{v}_{\text{est}} = \boldsymbol{x}_{\text{est}} - \boldsymbol{x}_{\text{est}-1}$.

\subsection{Systematic Search}

\begin{figure}[tbp]
\centering
\includegraphics[width=0.43\textwidth]{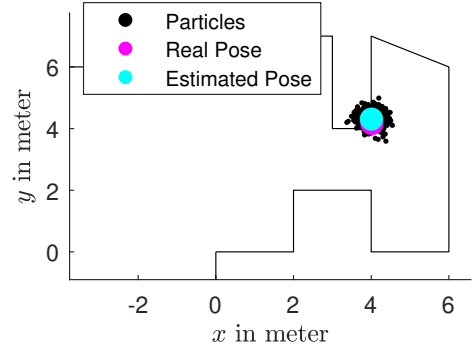}
\caption{Sampled Particles: Particles have been sampled around the first pose estimate generated using land navigation.}
\label{fig:ParticleSpreading}
\end{figure}

Using the land navigation from above we now have a first estimate on where the robot is located. Based on this estimate we are able to start a systematic search in this region to determine a more accurate and certain pose estimate. Therefore, we use a particle filter since other localization techniques, such as Kalman Filter, can not handle binary measurement data. The general idea of the particle filter \cite{thrun2005probabilistic} is to represent the probability distribution of the posterior by a set of samples, called particles, instead of using a parametric form as the Kalman Filter does. Here each particle
\begin{equation}
\mathcal{X}_t = \boldsymbol{x}_t^{[1]},\boldsymbol{x}_t^{[2]},\dots,\boldsymbol{x}_t^{[N]}
\end{equation} 
represents a concrete instantiation of the state at time $t$, where $N$ denotes the number of particles used. The belief $bel(\boldsymbol{x}_t)$ is then approximated by the set of particles $\mathcal{X}_t$. The Bayes filter posterior is used to include the likelihood of a state hypothesis $\boldsymbol{x}_t$
\begin{equation}
\boldsymbol{x}_t^{[i]} \sim p(\boldsymbol{x}_t | \boldsymbol{z}_{1:t},\boldsymbol{u}_{1:t}) .
\end{equation}
Here $\boldsymbol{z}_{1:t}$ and $\boldsymbol{u}_{1:t}$ represent the measurement history and the input signal history respectively. \\
The idea of our method is to reduce the particles required for global localization with a particle filter extensively by just sampling particles in an area around the position estimate retrieved by land navigation as shown above. This is nothing else than the transformation from a global localization problem to a local one. Therefore, we simply use a Gaussian distribution ${\mathcal{N}(\boldsymbol{\mu}, \boldsymbol{\Sigma})}$ with the mean ${\boldsymbol{\mu} = [\boldsymbol{x}_{\text{est}}^T, \varphi_{\text{est}}]^T}$ and parametrized covariance matrix ${\boldsymbol{\Sigma} = \text{diag}[\sigma_x, \sigma_y, \sigma_{\varphi}]}$. The parameters ${\sigma_x, \sigma_y,\sigma_{\varphi}}$ can be chosen accordingly to the mean and uncertainty of the error of the initial pose estimate as we will see in Section \ref{se:Results}. In Figure \ref{fig:ParticleSpreading}
an example distribution of the particles around the estimated pose is shown. After sampling the particles we now are able to further improve the localization executing a systematic search along the border line using the particle filter. Thereby, the weighting of the particles happens as follows: If a particle would see the same as the sensor has measured, then we give this particle the weight $w_i = \hat{w}$, where $w_i$ is the weight of the i-th particle. If the particle would not see the same, then $w_i = (1 - \hat{w})$. Here the parameter $\hat{w}$ has to be larger then $0.5$.

%

\section{\uppercase{Results}}\label{se:Results}

\noindent In this section, we evaluate the proposed algorithms in regard to stability and performance. Stability is defined as the fraction of localization experiments were the robot could localize itself from an initially lost pose in the presence of sensor noise. For that the Euclidian distance between the true and the estimated pose had to be below a threshold of $\SI{0.3}{\meter}$. Performance is defined as the mean accuracy of the pose estimation and the time required to generate the pose estimate. Therefore, we created a simulation environment using Matlab with a velocity and an odometry motion model for a differential drive robot as presented in \cite{thrun2005probabilistic}. For realistic conditions, we calibrated the noise parameters in experiments to match the true motion model of a real Viking MI 422P, a purchasable autonomous lawn mower. For this calibration, we tracked the lawn mower movements using a visual tracking system (OptiTrack) and computed the transition model parameters through maximum likelihood estimation. The parameters can be found in \tabref{tab:ParamModels}. We used a sampling frequency of $f = 20 \, Hz$ for the simulated system and the maximum linear velocity $v_0 = \SI{0.3}{\meter\per\second}$ and angular velocity $\omega = \SI{0.6}{\per\second}$. These values correspond approximately to the standard for purchasable low-cost robots. We set the number of $K = 100$ which leads to a period time of the wiggly lines $T = \SI{5}{\second}$. This is a trade-off between the enlargement of the scanning surface and the movement that can be performed on a real robot. The relative position of the binary sensor in regard to the robot frame, \eqref{eq:SensorPosition}, is given by $[\Delta x, \Delta y]^T = [\SI{0.3}{\meter},\SI{0.0}{\meter}]^T$.

\begin{table}[bt]
\centering
\caption{Measured parameters for the velocity and odometry motion model from \cite{thrun2005probabilistic}.} 
\label{tab:ParamModels}
\begin{tabular}{|c|c|c|}
\hline 
 & Vel. Motion Model & Odom. Motion Model \\ 
\hline 
$\alpha_1$ & 0.0346 & 0.0849 \\ 
\hline 
$\alpha_2$ & 0.0316 & 0.0412 \\ 
\hline 
$\alpha_3$ & 0.0755 & 0.0316 \\ 
\hline 
$\alpha_4$ & 0.0566 & 0.0173 \\ 
\hline 
$\alpha_5$ & 0.0592 & - \\ 
\hline 
$\alpha_6$ & 0.0678 & - \\ 
\hline 
\end{tabular}
\end{table}

\subsection{Wall Follower}

\begin{figure}[bp]
\centering
\includegraphics[width=0.43\textwidth]{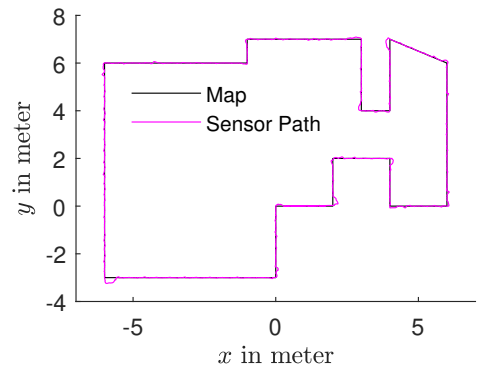}
\caption{Example of a path along the boundary line generated using the presented wall following algorithm with $a_{\mu} = 0.7$, $a_v = 0.7$, $M = 100$ and a sensor noise of $\SI{10}{\percent}$.}
\label{fig:ExampleWallFollowing}
\end{figure}

\begin{figure}[hbtp]
{
\centering
\includegraphics[width=0.43\textwidth]{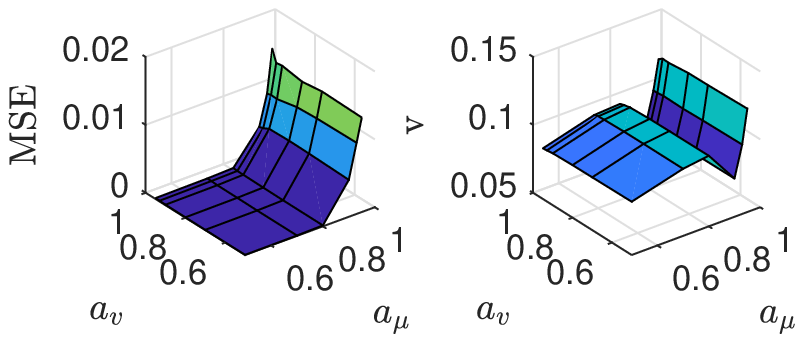}
\captionof{figure}{MSE and mean velocity for a noise factor of $\SI{0}{\percent}$ and varying parameters $a_{\mu}$ and $a_v$. The results shown are the average values over 10 iterations.}
\label{fig:MSE_V_00}
\medskip
\includegraphics[width=0.43\textwidth]{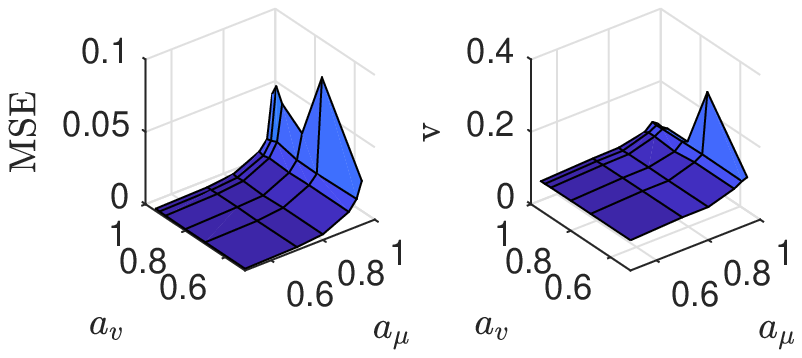}
\captionof{figure}{MSE and mean velocity for a noise factor of $\SI{20}{\percent}$ and varying parameters $a_{\mu}$ and $a_v$. The results shown are the average values over 10 iterations.}
\label{fig:MSE_V_02}
\medskip
\includegraphics[width=0.43\textwidth]{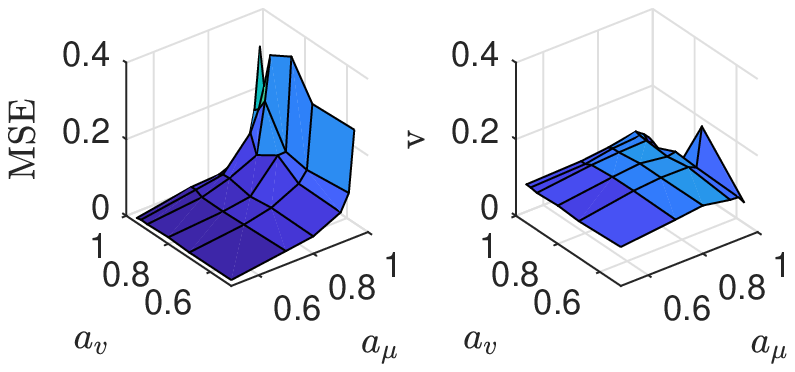}
\captionof{figure}{MSE and mean velocity for a noise factor of $\SI{40}{\percent}$ and varying parameters $a_{\mu}$ and $a_v$. The results shown are the average values over 10 iterations.}
\label{fig:MSE_V_04}
}
\end{figure}

\noindent As mentioned before, only at the boundary we are able to generate data useful for localization since only there we can detect differences in the signal generated by the given binary sensor. Hence, we first evaluate the performance of the presented wall following algorithm. Therefore, we use the mean squared error (MSE) to measure the deviation of the path of the sensor to the given boundary line of the map. An example of such a path along the map is given in \figref{fig:ExampleWallFollowing}. Let $\boldsymbol{x}_i$ be the i-th position of the sensor and $\boldsymbol{s}_i$ the closest point at the boundary line to $\boldsymbol{x}_i$, then
\begin{equation}
MSE = \frac{1}{N} \sum_{i=1}^{N} ||\boldsymbol{s}_i -  \boldsymbol{x}_i||^2 ,
\end{equation}
where $N$ represents the number of time steps required to complete one round along the boundary line. We also evaluate the mean velocity of the robot driving along the boundary
\begin{equation}
v_{\mu} = \frac{U}{T},
\end{equation}
where $U$ is the circumference of the map and $T$ the time required by the robot to complete one round. In \figref{fig:MSE_V_00}, \figref{fig:MSE_V_02} and \figref{fig:MSE_V_04}, the average simulation results over 10 simulations for different noise factors in \eqref{eq:amu} and \eqref{eq:av} are presented. The used map is shown in \figref{fig:ExampleWallFollowing}. The noise factor hereby determines the randomness of the binary signals, thus a factor of $0$ leads to always accurate measurements whereas a factor of $1$ implies total random measurement results. Hence, a noise factor of $\SI{40}{\percent}$ signals that $\SI{40}{\percent}$ of the output signals of the binary sensor are random.\\

\noindent The presented wall follower leads to a stable and accurate wall following behavior as  	exemplarily shown in \figref{fig:ExampleWallFollowing}. With increasing noise factor, the MSE becomes larger. However, the algorithm is stable enough to steer the robot along the boundary line even by noise factors around $\SI{40}{\percent}$. The algorithm seems to work best, thus leading to small MSE, for small $a_{\mu}$. However, a small value for $a_{\mu}$ leads to large changes in the linear and angular velocities which may not be feasible on a real robot or may cause large odometry erros. We discuss this problem more intensively in Section \ref{sec:Conclusion}. The parameter $a_v$ seems to have no strong influence neither on the MSE nor on the mean velocity.


\subsection{Land Navigation}\label{subse:ShapeComp}

\begin{figure}[bp]
\centering
\includegraphics[width=0.43\textwidth]{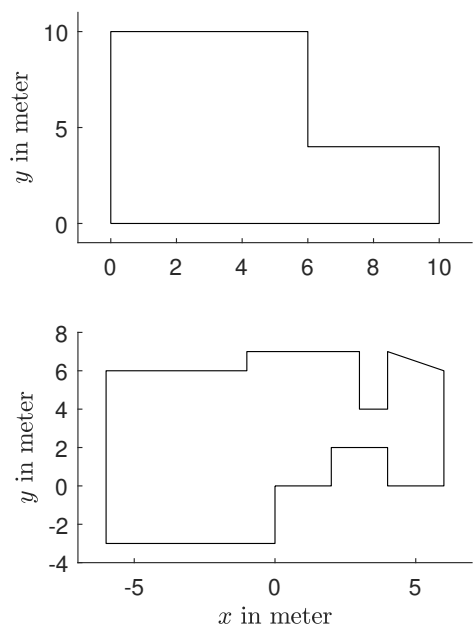}
\caption{Maps used for the simulation. Map 1 is at the top and map 2 at the bottom. The circumferences for the maps are $U_1 = \SI{40}{\meter}$ and $U_2 = \SI{53.23}{\meter}$.}
\label{fig:MapsSimulation}
\end{figure}

\begin{figure}[hbtp]
{
\centering
\includegraphics[width=0.43\textwidth]{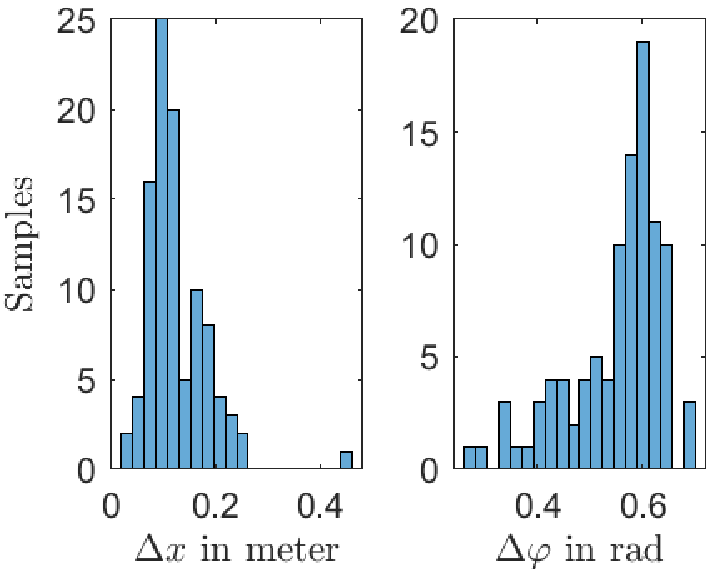}
\captionof{figure}{Evaluation results for map 1: Histogram of the difference between position and orientation estimate and true position and orientation after matching through shape comparison with ${\mu_{\Delta x} = 0.13}$, ${\mu_{\Delta \varphi} = 0.55}$, ${\sigma_{\Delta x} = 0.06}$, ${\sigma_{\Delta \varphi} = 0.09}$.}
\label{fig:HistMap1}
\medskip
\includegraphics[width=0.43\textwidth]{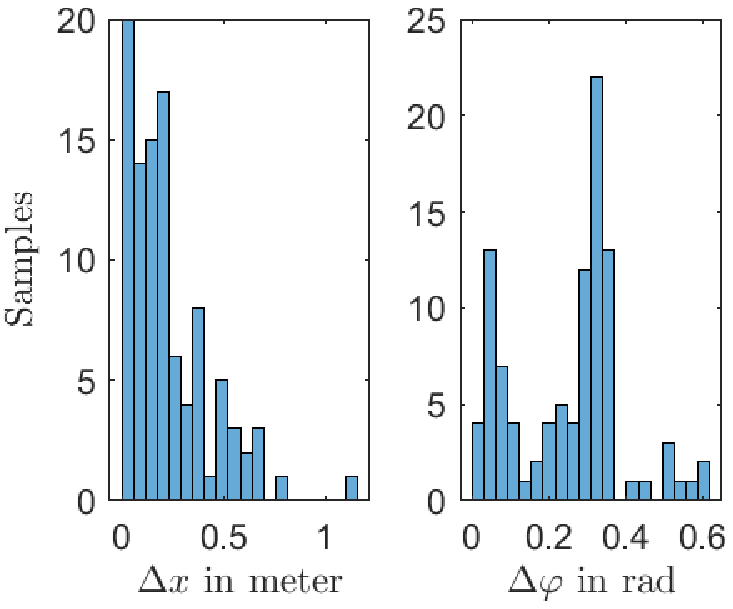}
\captionof{figure}{Evaluation results for map 2: Histogram of the difference between position and orientation estimate and true position and orientation after matching through shape comparison with ${\mu_{\Delta x} = 0.23}$, ${\mu_{\Delta \varphi} = 0.25}$, ${\sigma_{\Delta x} = 0.20}$, ${\sigma_{\Delta \varphi} = 0.15}$. }
\label{fig:HistMap2}
}
\end{figure}

In order to evaluate the performance and stability of the proposed shape comparison method we simulated the differential drive robot $100$ times in two different maps (\figref{fig:MapsSimulation}) with randomly chosen starting positions. We trained the parameters specifically for the given maps under usage of the proposed wall following algorithm with ${a_{\mu} = 0.7}$, ${a_v = 0.7}$, ${M = 100}$ and a simulated sensor noise of $\SI{10}{\percent}$. The resulting parameters are presented in \tabref{tab:TrainedParameters}.
\begin{table}[tb]
\centering
\caption{Trained parameters for the presented maps from \figref{fig:MapsSimulation}.}
\label{tab:TrainedParameters}
\begin{tabular}{c|c|c|c|c}
 & $L_{\text{min}}$ & $e_{\text{max}}$ & $c_{\text{min}}$ & $U_{\text{min}}$ \\ 
\hline 
Map 1 & $0.5$ & $0.01$ & $0.2$ & $0.5$ \\ 
\hline 
Map 2 & $0.5$ & $0.01$ & $0.3$ & $0.4$ \\ 
\end{tabular}
\end{table}
We then calculated the difference of the estimated position and the true position ${\Delta x = ||\boldsymbol{x}_{\text{true}} - \boldsymbol{x}_{\text{est}}||}$ as well as for the estimated orientation and true orientation ${\Delta \varphi = ||\varphi_{\text{true}} - \varphi_{\text{est}}||}$. Histograms of these errors are depicted in \figref{fig:HistMap1} and \figref{fig:HistMap2}. The mean time for finding this first pose estimate is around $T_1 = \SI{336}{\second}$ for map $1$ and $T_2 = \SI{382}{\second}$ for map $2$. The results show that the proposed algorithm is able to compute accurate estimates of the robots poses. Those estimates can then be further used to sample particles for a particle filter to systematically search in the region of interest for the correct pose.

\subsection{Systematic Search}

\begin{figure}[hbtp]
{
\centering
\includegraphics[width=0.43\textwidth]{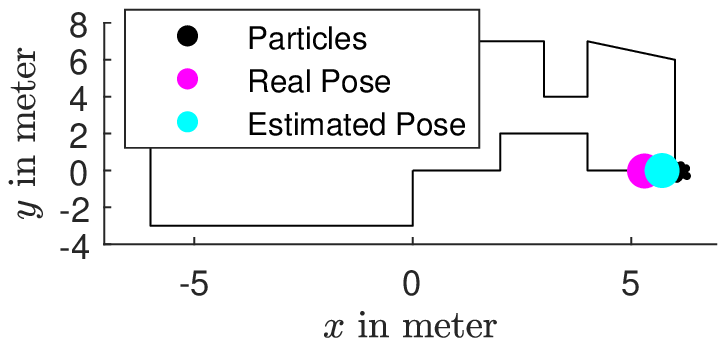}
\captionof{figure}{Randomly sampled particles around the position estimate $\boldsymbol{x}_{\text{est}}$.}
\label{fig:Particles01}
\medskip
\includegraphics[width=0.43\textwidth]{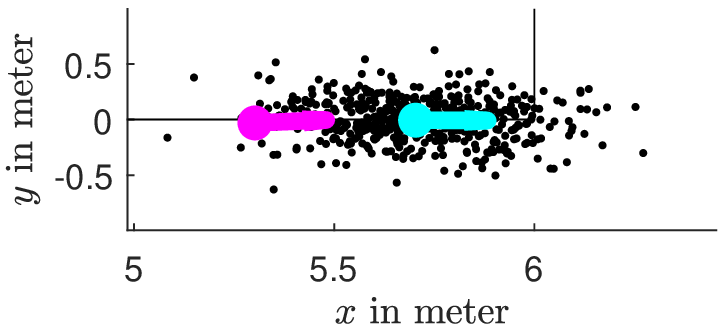}
\captionof{figure}{Randomly sampled particles around the position estimate $\boldsymbol{x}_{\text{est}}$ in close up view.}
\label{fig:Particles02}
\medskip
\includegraphics[width=0.43\textwidth]{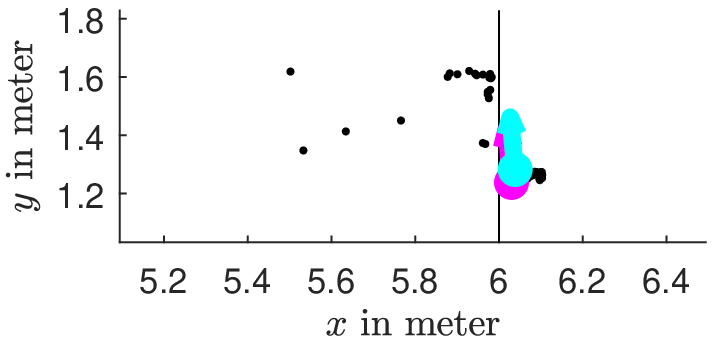}
\captionof{figure}{Pose estimate of the robot after the systematic search.}
\label{fig:Particles03}
\medskip
\includegraphics[width=0.43\textwidth]{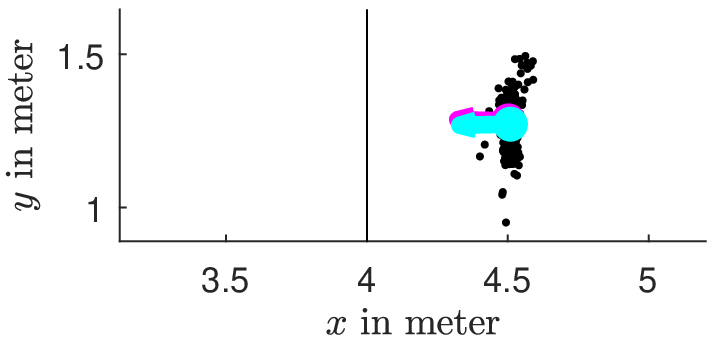}
\captionof{figure}{Moving within the map increases the uncertainty of the pose estimate.}
\label{fig:Particles04}
\medskip
\includegraphics[width=0.43\textwidth]{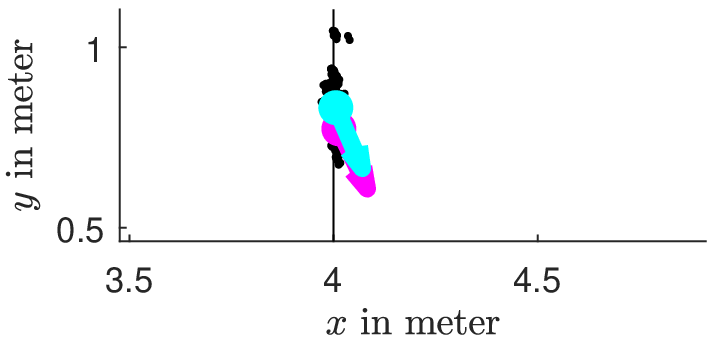}
\captionof{figure}{New information collected at the boundary leads to an improvement in pose estimation.}
\label{fig:Particles05}
}
\end{figure}

We now demonstrate how the position estimate from above can be further used to systematically search for a better estimate and for executing planning tasks. As proposed, we first sample particles around the position estimate using Gaussian distributions
\begin{equation}
\label{eq:SamplingParticles}
\begin{bmatrix}
x \\
y \\
\varphi
\end{bmatrix} \sim
\begin{bmatrix}
\mathcal{N}(x_{\text{est}}, \mu_{\Delta x}+3\sigma_{\Delta x}) \\
\mathcal{N}(y_{\text{est}}, \mu_{\Delta x}+3\sigma_{\Delta x}) \\
\mathcal{N}(\varphi_{\text{est}}, \mu_{\Delta \varphi}+3\sigma_{\Delta \varphi})
\end{bmatrix} .
\end{equation}
The standard deviation $\mu + 3 \sigma$, depending on the measurement results from Section \ref{subse:ShapeComp}, ensures that the particles are sampled in an area large enough around the initial pose estimate. In \figref{fig:Particles01} and \figref{fig:Particles02}, the particles and the pose estimate are shown as well as the true position of the robot. In the following step we systematically search for the true pose of the robot by following the boundary line. The particle filter algorithm weights the particles and does a resampling if required. We stop the wall following if we are certain enough about the pose of the robot, hence the variance of the particles is less then a certain treshold. In \figref{fig:Particles03} the robot after the systematic search for a better pose estimate is shown. Now we are able to execute certain tasks with the robot, for example mowing the lawn. Thereby, the robot has to move away from the boundary and the pose estimate becomes more uncertain as shown in \figref{fig:Particles04}. After reaching again the boundary line the particle filter is able to enhance the pose estimate based on the new information as depicted in \figref{fig:Particles05}. \\

\begin{figure}[bt]
\centering
\includegraphics[width=0.43\textwidth]{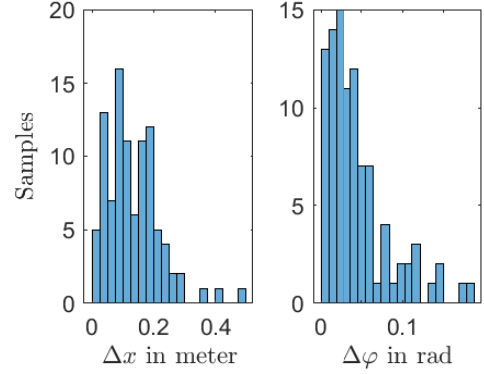}
\caption{Histogram of the difference between position and orientation estimate and true position and orientation after the systematic search using the particle filter with ${\mu_{\Delta x} = 0.13}$, ${\mu_{\Delta \varphi} = 0.04}$, ${\sigma_{\Delta x} = 0.086}$, ${\sigma_{\Delta \varphi} = 0.04}$.}
\label{fig:ParticleEvaluation}
\end{figure}

\noindent We also evaluated the performance of the proposed systematic search. Therefore, we simulated the differential drive robot $100$ times sampling the particles as proposed in \eqref{eq:SamplingParticles}. We then followed the boundary line until the standard deviation of the orientation has been below $0.2$. Again, we evaluated the difference between position and orientation estimate and true position and orientation. Here, 3 times the particle filter has not found a sufficient accurate pose estimate. If this happens, we simply can restart the procedure by finding a pose estimate using land navigation. In \figref{fig:ParticleEvaluation} a histogram depicting the results are shown. The systematic search for the pose of the robot using a particle filter improves the orientation estimate compared to the initial estimate presented in section \ref{subse:ShapeComp}.

%
%
%

\section{\uppercase{Conclusion}}\label{sec:Conclusion}

\noindent We have presented a method for global localization based only on odometry data and one binary sensor. This method, inspired by insect localization strategies, is essential for efficient path planning methods as presented in \cite{hess2014probabilistic} or \cite{galceran2013survey} which require an accurate estimate of the robots pose. Our approach combines ideas of path integration, land navigation and systematic search to find a pose estimate for the robot.\\
Our method relies on a stable and accurate wall following algorithm. We showed that the proposed algorithm is robust even under $\SI{40}{\percent}$ sensor noise and, if convenient parameters are chosen, leads to an accurate wall following behavior.\\
The proposed approach for finding a first pose estimate is robust and accurate. Given this pose estimate we are able to sample particles around this position for starting a systematic search with a particle filter algorithm. Therefore, we maintain the wall following behavior until the uncertainty of the pose estimate of the particle filter is below a certain threshold. The proposed search method improves the orientation estimate intensively while maintaining the accuracy of the position estimate. As mentioned in the Introduction, it is not possible to compare our approach directly with other approaches since different types of sensors are used. However, comparing our localization results with the results presented in \cite{stavrou2012localization} for a particle filter with a single low-range sensor, we reach a similar accuracy for the position estimate by only using a binary sensor. Thereby, the accuracy is in the range of around $\SI{0.1}{\meter}$. \\ 
Using the generated pose estimate we are now able to execute given tasks by constantly relocating the robot at the boundary lines. For example, using bioinspired neural networks for complete coverage path planning \cite{zhang2017discrete}. Open research questions in regard to the proposed approach are which methods can be used to learn the algorithm parameters $L_{\text{min}}$, $e_{\text{max}}$, $c_{\text{min}}$, $U_{\text{min}}$ on the fly. Also, the algorithm has to be evaluated on a real robotic system in a realistic garden setup.\\ 

\balance

\bibliographystyle{apalike}
{\small
\bibliography{PaperLocalization}}

\end{document}